\documentclass[10pt,twocolumn,letterpaper]{article}
\usepackage[pagenumbers]{cvpr} % To force page numbers, e.g. for an arXiv version

\definecolor{cvprblue}{rgb}{0.21,0.49,0.74}
\usepackage[pagebackref,breaklinks,colorlinks,allcolors=cvprblue]{hyperref}

\usepackage{multirow}
\usepackage{array}
\usepackage{adjustbox}
\usepackage{graphicx}   % 用于插入图片
\usepackage{subcaption} % 用于子图
\def\confName{CVPR}
\def\confYear{2026}

\title{Qwen-Image-Layered: Towards Inherent Editability via Layer Decomposition}

\author{
Shengming Yin$^{1}$ \quad Zekai Zhang$^{2}$ \quad Zecheng Tang$^{2}$ \quad Kaiyuan Gao$^{2}$ \\
Xiao Xu$^{2}$ \quad Kun Yan$^{2}$ \quad Jiahao Li$^{2}$  \quad Yilei Chen$^{2}$ \quad Yuxiang Chen$^{2}$ \\
Heung-Yeung Shum$^{3}$ \quad  Lionel M. Ni$^{1}$ \quad Jingren Zhou$^{2}$ \quad Junyang Lin$^{2}$ \quad Chenfei Wu$^{2}$\thanks{\scriptsize Corresponding author.} \\
{$^{1}$HKUST(GZ)\quad $^{2}$Alibaba
 \quad $^{3}$HKUST} \\
}

\begin{document}
\makeatletter
\let\@oldmaketitle\@maketitle
\renewcommand{\@maketitle}{
  \@oldmaketitle
  \begin{minipage}{\linewidth}
    \vspace{-8mm}
    \includegraphics[width=\linewidth]{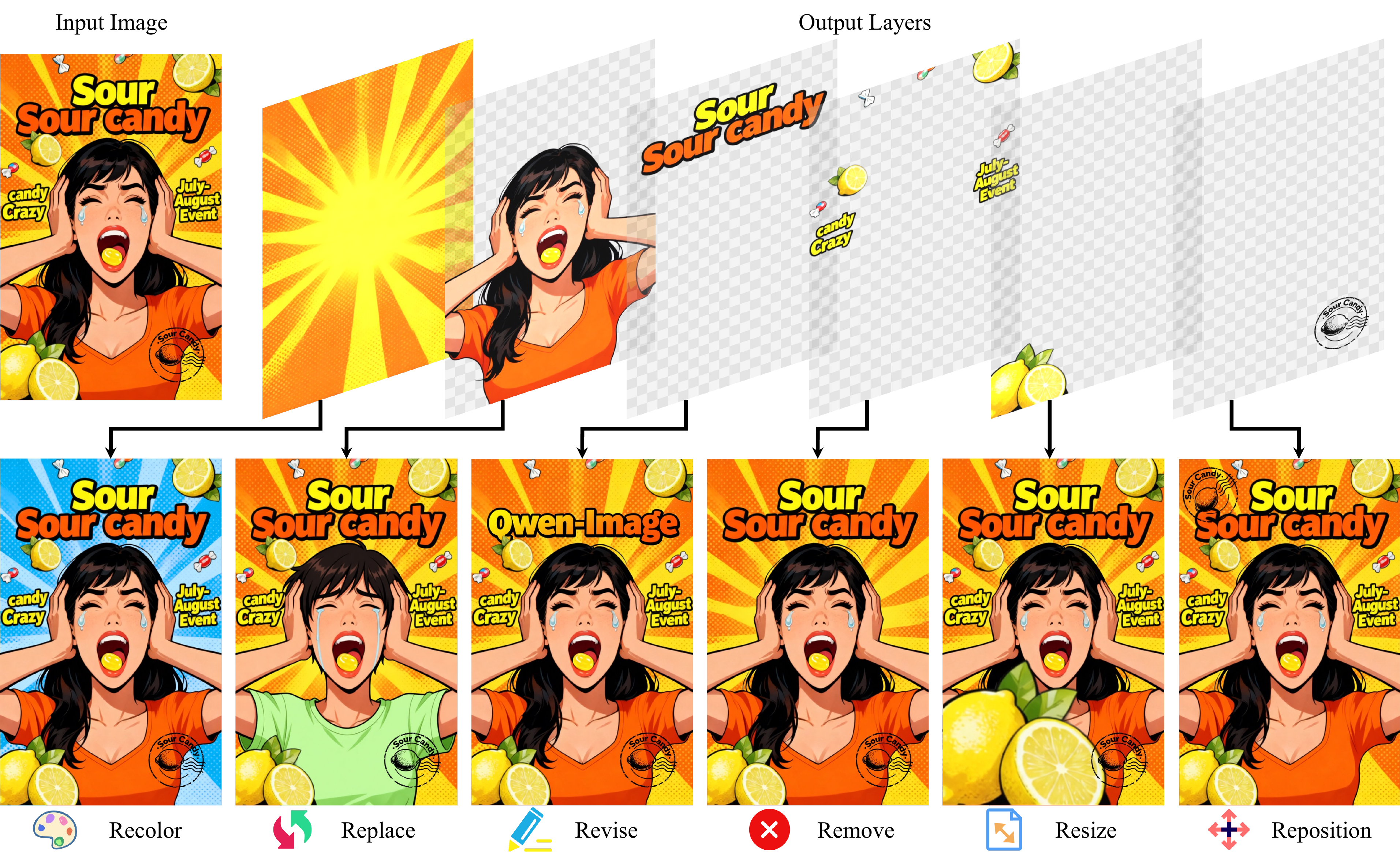} 
    \vspace{-8mm}
    \captionof{figure}{Qwen-Image-Layered is capable of decomposing an input image into multiple semantically disentangled RGBA layers, thereby enabling inherent editability, where each layer can be independently manipulated without affecting other content.}
    \vspace{2mm}
    \label{fig:teaser}
  \end{minipage}
  }
\makeatother

\maketitle
\begin{abstract}

\vspace{-4mm}
Recent visual generative models often struggle with consistency during image editing due to the entangled nature of raster images, where all visual content is fused into a single canvas. In contrast, professional design tools employ layered representations, allowing isolated edits while preserving consistency. Motivated by this, we propose \textbf{Qwen-Image-Layered}, an end-to-end diffusion model that decomposes a single RGB image into multiple semantically disentangled RGBA layers, enabling \textbf{inherent editability}, where each RGBA layer can be independently manipulated without affecting other content. To support variable-length decomposition, we introduce three key components: (1) an RGBA-VAE to unify the latent representations of RGB and RGBA images; (2) a VLD-MMDiT (Variable Layers Decomposition MMDiT) architecture capable of decomposing a variable number of image layers; and (3) a Multi-stage Training strategy to adapt a pretrained image generation model into a multilayer image decomposer. Furthermore, to address the scarcity of high-quality multilayer training images, we build a pipeline to extract and annotate multilayer images from Photoshop documents (PSD). Experiments demonstrate that our method significantly surpasses existing approaches in decomposition quality and establishes a new paradigm for consistent image editing. Our code and models are released on \href{https://github.com/QwenLM/Qwen-Image-Layered}{https://github.com/QwenLM/Qwen-Image-Layered}

\begin{figure*}[p]
    \centering
    \includegraphics[width=\linewidth]{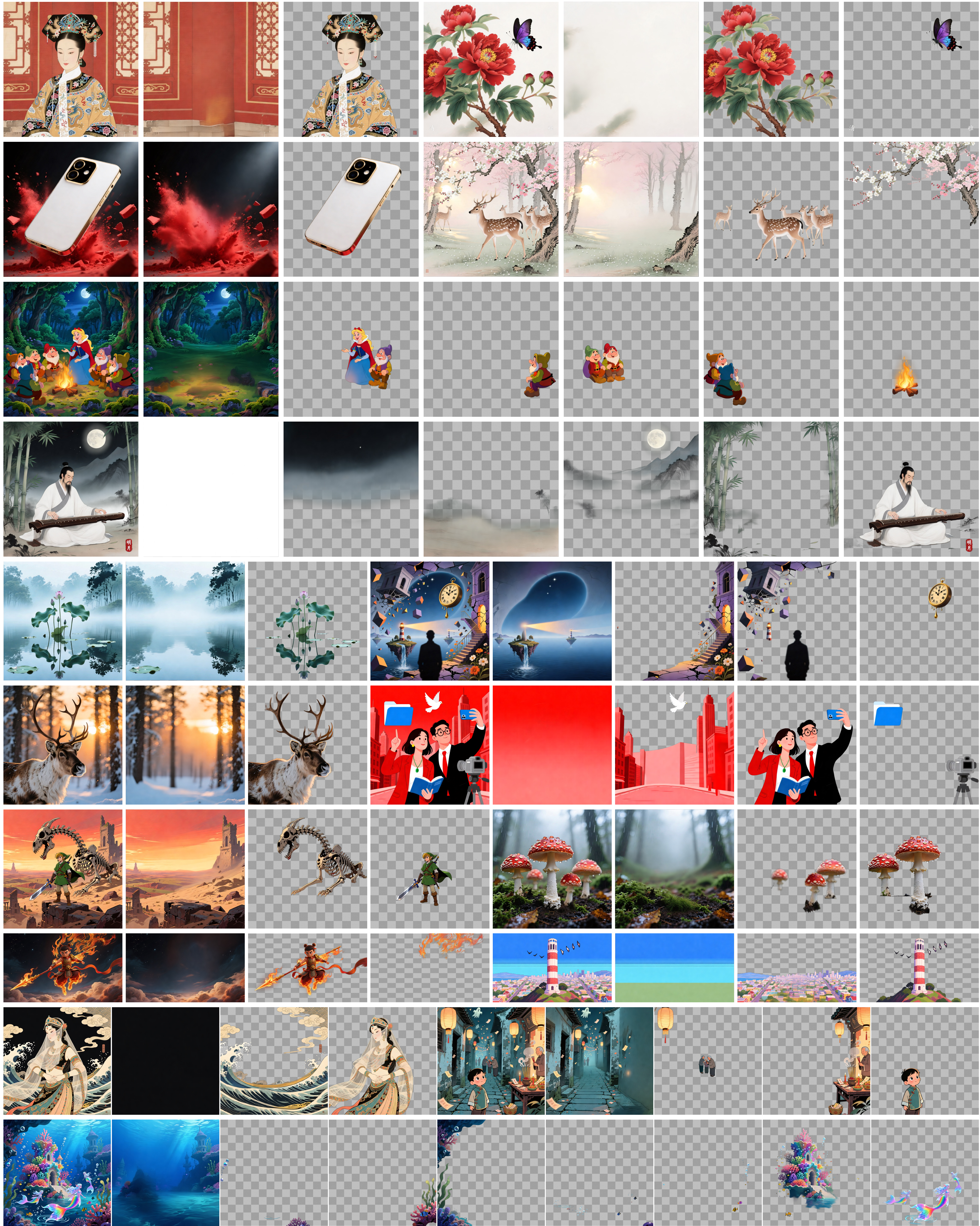}
    \vspace{-6mm}
    \caption{Visualization of Image-to-Multi-RGBA (I2L) on open-domain images. The leftmost column in each group shows the input image. Qwen-Image-Layered is capable of decomposing diverse images into high-quality, semantically disentangled layers, where each layer can be independently manipulated without affect other content.}
    \label{showcase:1}
\end{figure*}

\begin{figure*}[p]
    \centering
    \includegraphics[width=\linewidth]{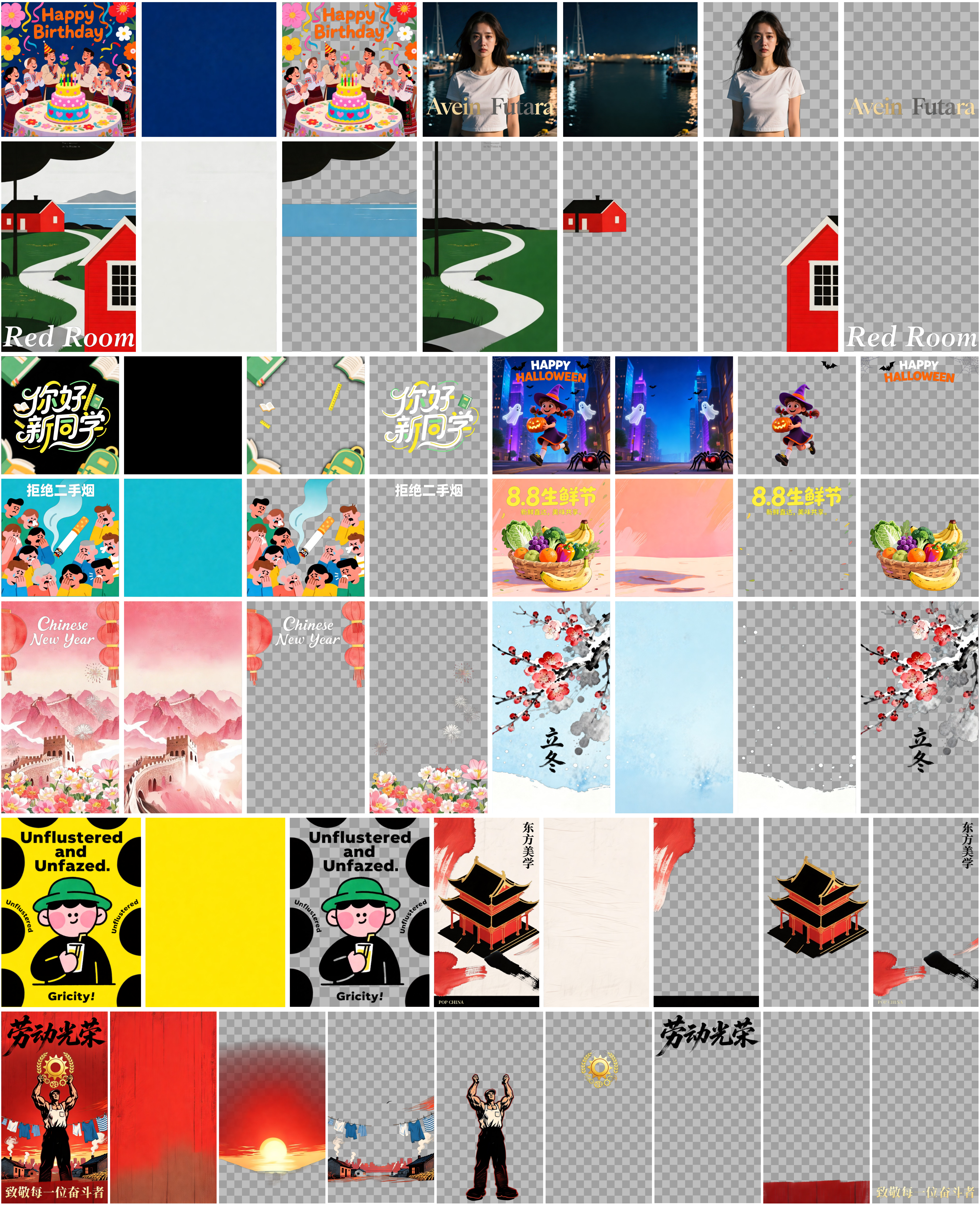}
    \vspace{-8mm}
    \caption{Visualization of Image-to-Multi-RGBA (I2L) on images containing texts. The leftmost column in each group shows the input image. Qwen-Image-Layered is capable of accurately decomposing both text and objects into semantically disentangled layers.}
    \label{showcase:2}
\end{figure*}

\end{abstract}

\vspace{-6mm}
\section{Introduction}\label{sec:intro}
Recent advances in visual generative models have enabled impressive image synthesis capabilities~\cite{wu2022nuwa,liang2022nuwa,rombach2022high,podell2023sdxl,esser2024scaling,gong2025seedream,gao2025seedream,cai2025hidream,wu2025qwen}. However, in the context of image editing, achieving precise modifications while preserving the structure and semantics of unedited regions remains a significant challenge. This issue typically appears as semantic drift (\eg unintended changes to a person's identity) and geometric misalignment (\eg shifts in object position or scale). 

Existing editing approaches fail to fundamentally address this problem. Global editing methods~\cite{brooks2023instructpix2pix,zhang2023magicbrush,wang2025seededit,deng2025bagel,labs2025kontext,wu2025omnigen2,liu2025step1x}, which resample the entire image in the latent space of generative models, are inherently limited by the stochastic nature of probabilistic generation and thus cannot ensure consistency in unedited regions. Meanwhile, mask-guided local editing methods~\cite{couairon2022diffedit,mao2024mag,simsar2025lime} restrict modification within user-specified masks. However, in complex scenes, especially those involving occlusion or soft boundaries, the actual editing region is often ambiguous, thus failing to fundamentally solve the consistency problem.

Rather than tackling this issue purely through model design or data engineering, we argue that the core challenge lies in the representation of images themselves. Traditional raster images are flat and entangled: all visual content is fused into a single canvas, with semantics and geometry tightly coupled. Consequently, any edit inevitably propagates through this entangled pixel space, leading to the aforementioned inconsistencies.

To overcome this fundamental limitation, we advocate for a naturally disentangled image representation. Specifically, we propose representing an image as a stack of semantically decomposed RGBA layers, as illustrated in the upper part of \cref{fig:teaser}. This layered structure enables inherent editability with built-in consistency: edits are applied exclusively to the target layer, physically isolating them from the rest of the content, and thereby eliminating semantic drift and geometric misalignment. Moreover, such a layer-wise representation naturally supports high-fidelity elementary operations—such as resizing, repositioning, and recoloring, as demonstrated in the lower part of \cref{fig:teaser}.

Based on this insight, we introduce Qwen-Image-Layered, an end-to-end diffusion model that directly decomposes a single RGB image into multiple semantically disentangled RGBA layers. Once decomposed, each layer can be independently manipulated while leaving all other content exactly unchanged—enabling truly consistent image editing. To support variable-length decomposition, our image decomposer is built upon three key designs: (1) an RGBA-VAE that establishes a shared latent space for both RGB and RGBA images; (2) a VLD-MMDiT (Variable Layers Decomposition MMDiT) architecture that enables training with a variable number of layers; and (3) a Multi-stage Training strategy that progressively adapts a pretrained image generation model into an multilayer image decomposer. Furthermore, to address the scarcity of high-quality multilayer image data, we develop a data pipeline to filter and annotate multilayer images from real-world Photoshop documents (PSD). 

We summarize our contributions as follows:
\begin{itemize}
    \item We propose \textbf{Qwen-Image-Layered}, an end-to-end diffusion model that decomposes an image into multiple high-quality, semantically disentangled RGBA layers, thereby enabling inherently consistent image editing.
    \item We design the image decomposer from three aspects: 1) an RGBA-VAE to provide shared latent space for RGB and RGBA images. 2) a VLD-MMDiT architecture to facilitate decomposition with variable number of layers. 3) a Multi-stage Training strategy to adapt a pretrained image generation model to a multilayer image decomposer.
    \item We develop a data processing pipeline to extract and annotate multilayer images from Photoshop documents, addressing the lack of high-quality multilayer images.
    \item Extensive experiments demonstrate that Qwen-Image-Layered not only outperforms existing methods in decomposition quality but also unlocks new possibilities for consistent, layer-based image editing and synthesis.
\end{itemize}

\section{Related Work}

\subsection{Image Editing}
Image editing has made significant progress in recent years and can be broadly categorized into two paradigms: global editing and mask-guided local editing. Global editing methods~\cite{brooks2023instructpix2pix,zhang2023magicbrush,wang2025seededit,deng2025bagel,labs2025kontext,wu2025omnigen2,wu2025qwen,liu2025step1x} regenerate the entire image to achieve holistic modifications, such as expression editing and style transfer. Among these, Qwen-Image-Edit~\cite{wu2025qwen} leverages two distinct yet complementary feature representations—semantic features from Qwen-VL~\cite{bai2025qwen2} and reconstructive features from VAE~\cite{kingma2013auto}—to enhance consistency. However, due to the inherent stochasticity of generative models, these approaches cannot ensure consistency in unedited regions. In contrast, mask-guided local editing methods~\cite{couairon2022diffedit,mao2024mag,simsar2025lime} constrain modifications within a specified mask to preserve global consistency. DiffEdit~\cite{couairon2022diffedit}, for instance, first automatically generates a mask to identify regions requiring modification and then edits the target area. Although intuitive, these approaches struggle with occlusions and soft boundaries, making it difficult to precisely identify the actual editing region and thus failing to fundamentally resolve the consistency issue.  Unlike these works, we propose decomposing the image into semantically disentangled RGBA layers, where each layer can be independently modified while keeping the others unchanged, thereby fundamentally ensuring consistent across edits.

\subsection{Image Decomposition}
Numerous studies have attempted to decompose images into  layers. Early approaches addressed this problem by performing segmentation in color space~\cite{tan2015decomposing,koyama2018decomposing,aksoy2017unmixing}. Subsequent work has focused on object-level decomposition in natural scenes~\cite{zhan2020self,monnier2021unsupervised,liu2024object}. Among these, PCNet~\cite{zhan2020self} learns to recover fractional object masks and contents in a self-supervised manner. More recent research has explored decomposing images into multiple RGBA layers~\cite{tudosiu2024mulan,yang2025generative,kang2025layeringdiff,suzuki2025layerd,chen2025rethinking}. One class of these methods leverages segmentation~\cite{ravi2024sam} or matting~\cite{li2024matting} to extract foreground objects, followed by image inpainting~\cite{yu2023inpaint} to reconstruct the background. For instance, LayerD~\cite{suzuki2025layerd} iteratively extracts the topmost unoccluded foreground layer and completes the background. Accordion~\cite{chen2025rethinking} proposes using Vision-Language Models~\cite{liu2023visual} to guide this decomposition process. Another category of work introduces mask-guided, object-centric image decomposition~\cite{yang2025generative,kang2025layeringdiff}, which decomposes an image into foreground and background layers based on a provided mask. These methods generally require segmentation to provide initial mask. However, segmentation often struggles with complex spatial layouts and the presence of multiple semi-transparent layers, resulting in low-quality layers. Moreover, multilayer decomposition typically requires recursive inference, leading to error propagation. Consequently, existing methods fail to produce complete, high-fidelity RGBA layers suitable for editing. In contrast to the aforementioned approaches, Qwen-Image-Layered employs an end-to-end framework to decompose input images directly into multiple high-quality RGBA layers, thereby enhancing decomposition quality and enabling consistency-preserving image editing.

\subsection{Multilayer Image Synthesis}

Multilayer image synthesis has also garnered sustained attention~\cite{zhang2023text2layer,zhang2024transparent,huang2024layerdiff,huang2025dreamlayer,pu2025art,chen2025prismlayers,huang2025psdiffusion,kang2025layeringdiff}. As a pioneer in layered image generation, Text2Layer~\cite{zhang2023text2layer} first trains a two-layer image autoencoder~\cite{kingma2013auto} and subsequently trains a diffusion model~\cite{ho2020denoising} on the latent representations, enabling the creation of two-layer images. LayerDiffusion~\cite{zhang2024transparent} introduces latent transparency into VAE and employs two different LoRA~\cite{hu2022lora} with shared attention to generate foreground and background. Through carefully designed inter-layer and intra-layer attention mechanisms, LayerDiff~\cite{huang2024layerdiff} is able to synthesize semantically consistent multilayer images. To achieve controllable multilayer image generation, ART~\cite{pu2025art} proposes an anonymous region layout to explicitly control the layout. LayeringDiff~\cite{kang2025layeringdiff} first generates a raster image using existing text-to-image models, and then decomposes it into foreground and background based on a mask. Qwen-Image-Layered is capable of decomposing AI-generated raster images into multiple RGBA layers, thus enabling multilayer image generation.
\section{Method}
\begin{figure*}[!ht]
    \centering
    \includegraphics[width=\linewidth]{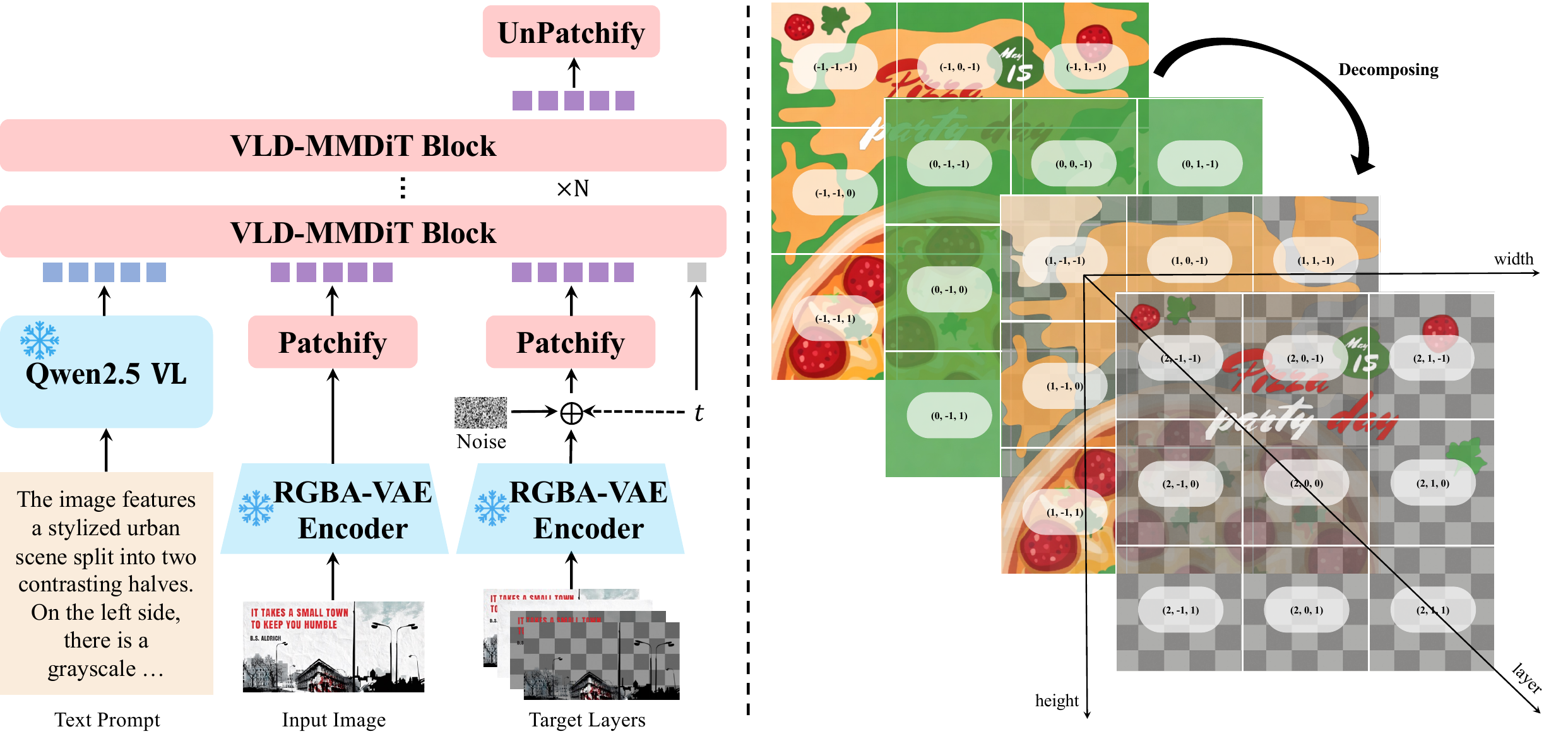}
    \vspace{-8mm}
    \caption{Overview of Qwen-Image-Layered. Left: Illustration of our proposed VLD-MMDiT (Variable Layers Decomposition MMDiT), where the input RGB image and the target RGBA layers are both encoded by our proposed RGBA-VAE. During attention computation, these two sequences are concatenated along the sequence dimension, thereby enhancing inter-layer and intra-layer interactions. Right: Illustration of Layer3D RoPE, where a new layer dimension is introduced to support a variable number of layers.}
    \vspace{-4mm}
    \label{fig:overview}
\end{figure*}

We propose an end-to-end layering approach that directly decomposes an input RGB image $I\in\mathbb{R}^{H\times W\times 3}$ into $N$ RGBA layers $L\in\mathbb{R}^{N\times H\times W\times 4}$, where each layer $L_i$ comprises a color component $RGB_i$ and an alpha matte $\alpha_i$, i.e. $L_i=[RGB_i;\alpha_i]$. The original image can be reconstructed by sequential alpha blending as follows:
\begin{align*}
    C_0&=\textbf{0} \\
    C_i&=\alpha_i\cdot RGB_i +(1-\alpha_i)\cdot C_{i-1} \quad \operatorname{i=1,...,N}
\end{align*}
where $C_i$ denotes the composite of the first $i$ layers, and the final composite satisfies $I=C_N$.
Building upon Qwen-Image~\cite{wu2025qwen}, we develop Qwen-Image-Layered from the following three aspects:

\begin{itemize}
    \item 1) In contrast to previous decomposer~\cite{yang2025generative} that employs separate VAEs, we propose an RGBA-VAE that encodes both RGB and RGBA images. This approach narrows the latent distribution gap between the input RGB image and the output RGBA layers.
    \item 2) Unlike prior methods that decompose images into foreground and background~\cite{kang2025layeringdiff,yang2025generative}, we propose a VLD-MMDiT (Variable Layers Decomposition MMDiT), which supports decomposition into a variable number of layers and is compatible with multi-task training.
    \item 3) To progressively adapt pretrained image generation model into a multilayer image decomposer, we design a multi-stage, multi-task training scheme that progressively evolves from simpler tasks to more complex ones.
\end{itemize}

\subsection{RGBA-VAE}

Variational Autoencoders (VAEs)~\cite{kingma2013auto} are commonly employed in diffusion models~\cite{rombach2022high} to reduce the dimensionality of the latent space, thereby improving both training and sampling efficiency.  In previous work, LayeringDiff~\cite{kang2025layeringdiff} utilized an RGB VAE to first generate the foreground layer and subsequently applied an additional module to obtain transparency. LayerDecomp~\cite{yang2025generative} adopted separate VAEs for the input RGB image and the output RGBA layers, resulting in a distribution gap between the input and output representations. To address these limitations, we propose RGBA VAE, a four-channel VAE designed to process both RGB and RGBA images.

Inspired by AlphaVAE~\cite{wang2025alphavae}, we extend the first convolution layer of the Qwen-Image VAE encoder $\mathcal{E}$ and the last convolution layer of the decoder $\mathcal{D}$ from three to four channels. To enable reconstruction of both RGB and RGBA images, we train it using both types of images. For RGB images, the alpha channel is set to 1. To maintain RGB reconstruction performance during initialization, we employ the following initialization strategy. Let $W_{\mathcal{E}}^0\in \mathbb{R}^{D_0\times 4\times k \times k \times k}$ and $b_{\mathcal{E}}^0\in \mathbb{R}^{D_0}$ denote the weight and bias of the first convolution layer in the encoder, and $W_{\mathcal{D}}^{l}\in \mathbb{R}^{4\times D_l\times k \times k \times k}$ and $b_{\mathcal{D}}^{l}\in \mathbb{R}^{4}$ denote  those of the last convolution layer in the decoder, where $k$ is the kernel size. We copy the parameters from the pretrained RGB VAE into the first three channels and set the newly initialized parameters as
\begin{align*}
    W_{\mathcal{E}}^0[:,3,:,:,:]=0 \quad W_{\mathcal{D}}^{l}[3,:,:,:,:]=0 \quad b_{\mathcal{D}}^{l}[3]=1 
\end{align*}
For the training objective, we use a combination of reconstruction loss, perceptual loss, and regularization loss. After training, both the input RGB image and the output RGBA layers are encoded into a shared latent space, where each RGBA layer is encoded independently. Notably, these layers exhibit no cross-layer redundancy; consequently, no compression is applied along the layer dimension.

\subsection{Variable Layers Decomposition MMDiT}\label{sec:mmdit}
Previous studies~\cite{yang2025generative,kang2025layeringdiff,suzuki2025layerd,chen2025rethinking} typically decompose images into background and foreground, requiring recursive inference to perform multilayer decomposition. Instead, Qwen-Image-Layered proposes VLD-MMDiT (Variable Layers Decomposition MMDiT) to facilitate the decomposition of a variable number of layers.

For Qwen-Image-Layered, it tasks an RGB image $I\in\mathbb{R}^{H\times W\times 3}$ as input and decomposes it into multiple RGBA layers $L\in\mathbb{R}^{N\times H\times W\times 4}$. Following Qwen-Image, we adopt the Flow Matching training objective. Formally, let $x_0\in \mathbb{R}^{N\times h\times w\times c}$ denote the latent representation of the target RGBA layers $L$, i.e., $x_0=\mathcal{E}(L)$. Then we sample noise $x_1$ from standard multivariate normal distribution and a timestep $t\in [0,1]$ from a logit-normal distribution. According to Rectified Flow~\cite{liu2022flow}, the intermediate state $x_t$ and velocity $v_t$ at timestep $t$ is defined as
\begin{align*}
    x_t &= tx_0+(1-t)x_1 \\
    v_t &= \frac{dx_t}{dt}=x_0-x_1
\end{align*}
For the input RGB image $I$, we also use RGBA-VAE to encode it as a latent representation $z_I\in\mathbb{R}^{h\times w\times c}$. Following Qwen-Image, the text prompt is encoded into text condition $h$ with MLLM. In practice, we can use Qwen2.5-VL~\cite{bai2025qwen2} to automatically generate the caption for the input image. Then, the model is trained to predict the target velocity with loss function defined as the mean squared error between the predicted velocity $v_\theta (x_t,t,z_I,h)$ and the ground truth $v_t$:
\begin{align*}
    \mathcal{L}=\mathbb{E}_{(x_0,x_1,t,z_I,h)\sim \mathcal{D}}||v_\theta (x_t,t,z_I,h)-v_t||^2
\end{align*}
where $\mathcal{D}$ denotes the training dataset.

Previous studies~\cite{huang2024layerdiff,huang2025dreamlayer} have achieved multilayer image generation through sophisticatedly designed inter-layer and intra-layer attention mechanisms. In contrast, we employ a Multi-Modal attention~\cite{esser2024scaling} to directly model these relationships, as shown in the left part of~\cref{fig:overview}. Specifically, we apply $2\times$ patchification to the noise-free input image $z_I$ and the intermediate state $x_t$ along the height and width dimensions. In each VLD-MMDiT block, two separate sets of parameters are used to process textual $h$ and visual information $z_I,x_t$ respectively. During attention computation, we concatenate these three sequences, thereby directly modeling both intra-layer and inter-layer interactions.

As shown in the right part of~\cref{fig:overview}, we propose a Layer3D RoPE within each VLD-MMDiT block to enable the decomposition of a variable number of layers, while supporting various tasks. Our design is inspired by the MSRoPE from Qwen-Image~\cite{wu2025qwen}, where the positional encoding in each layer is shifted towards the center. To accommodate a variable number of layers, we introduce an additional layer dimension. For the intermediate state $x_t$, the layer index starts from 0, and increases accordingly. For conditional image input $z_I$, we assign a layer index of -1, ensuring a clear distinction from any positive layer indices used in other tasks, \eg text-to-multilayer image generation.

\subsection{Multi-stage Training}
Directly finetuning a pretrained image generation model to perform image decomposition poses significant challenges, as it not only requires adapting to a new VAE but also involves learning new tasks. To address this issue, we propose a multi-stage, multi-task training scheme that progressively evolves from simpler tasks to more complex ones.

\textbf{Stage 1: From Text-to-RGB to Text-to-RGBA.}
We begin by adapting MMDiT to the latent space of RGBA VAE. At this stage, we replace the original VAE and train the model jointly on both text-to-RGB and text-to-RGBA generation tasks. This enables the model to generate not only standard raster images (RGB) but also images with transparency (RGBA).

\textbf{Stage 2: From Text-to-RGBA to Text-to-Multi-RGBA.}
Initially, the image generator is capable of producing only a single image. To support multilayer generation and adapt to the newly initialized layer dimension, we introduce a text-to-multiple-RGBA generation task. Following ART~\cite{pu2025art}, the model is trained to jointly predict both the final composite image and its corresponding transparent layers, thereby facilitating information propagation between the composite image and its layers. We refer to this model as Qwen-Image-Layered-T2L.

\textbf{Stage 3: From Text-to-Multi-RGBA to Image-to-Multi-RGBA.}
Up to this point, all tasks have been conditioned exclusively on textual prompts. In this stage, we introduce an additional image input, as detailed in \cref{sec:mmdit}, extending the model’s capability to decompose a given RGB image into multiple RGBA layers. We refer to this model as Qwen-Image-Layered-I2L.

\section{Experiment}
\subsection{Data Collection and Annotation}
\begin{figure}[t]
    \centering
    \begin{subfigure}[b]{0.24\textwidth}
        \includegraphics[width=\textwidth]{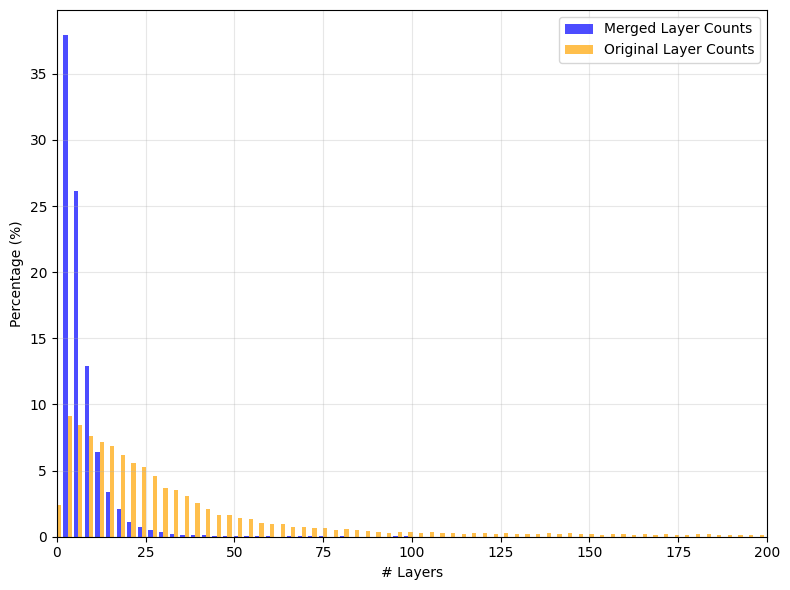} 
        \caption{Distribution of Layer Counts}
        \label{fig:numberoflayer}
    \end{subfigure}
    \begin{subfigure}[b]{0.22\textwidth}
        \includegraphics[width=\textwidth]{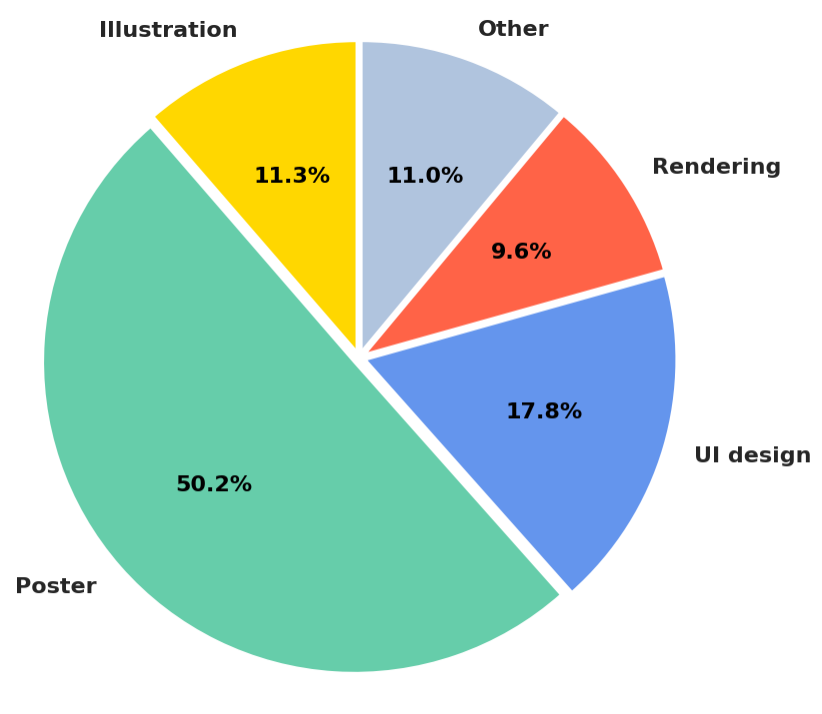} 
        \caption{Category Distribution}
        \label{fig:types}
    \end{subfigure}
    \vspace{-2mm}
    \caption{Statistics of the processed multilayer image dataset. (a) Distribution of layer counts before and after merging. (b) Category distribution in the final dataset. }
    \label{fig:main}
    \vspace{-4mm}
\end{figure}

Due to the scarcity of high-quality multilayer images, previous studies~\cite{zhang2023text2layer,huang2024layerdiff,kang2025layeringdiff,suzuki2025layerd} have largely relied on either synthetic data~\cite{tudosiu2024mulan} or simple graphic design datasets (\eg, Crello~\cite{yamaguchi2021canvasvae}), which typically lack complex layouts or semi-transparent layers. To bridge this gap, we developed a data pipeline to filter and annotate multilayer images derived from real world PSD (Photoshop Document) files.

We began by collecting a large corpus of PSD files and extracting all layers using \texttt{psd-tools}, an open-source Python library for parsing Adobe Photoshop documents. To ensure data quality, we filtered out layers containing anomalous elements, such as blurred faces. To improve decomposition performance, we removed non-contributing layers that do not influence the final composite image. Furthermore, given that some PSD files contain hundreds of layers—thereby increasing model complexity—we merged spatially non-overlapping layers to reduce the total layer count. As shown in \cref{fig:numberoflayer}, this operation substantially reduces the number of layers. Finally, we employed Qwen2.5-VL~\cite{bai2025qwen2} to generate text descriptions for the composite images, enabling Text-to-Multi-RGBA generation.

\begin{figure*}[!ht]
    \centering
    \includegraphics[width=\linewidth]{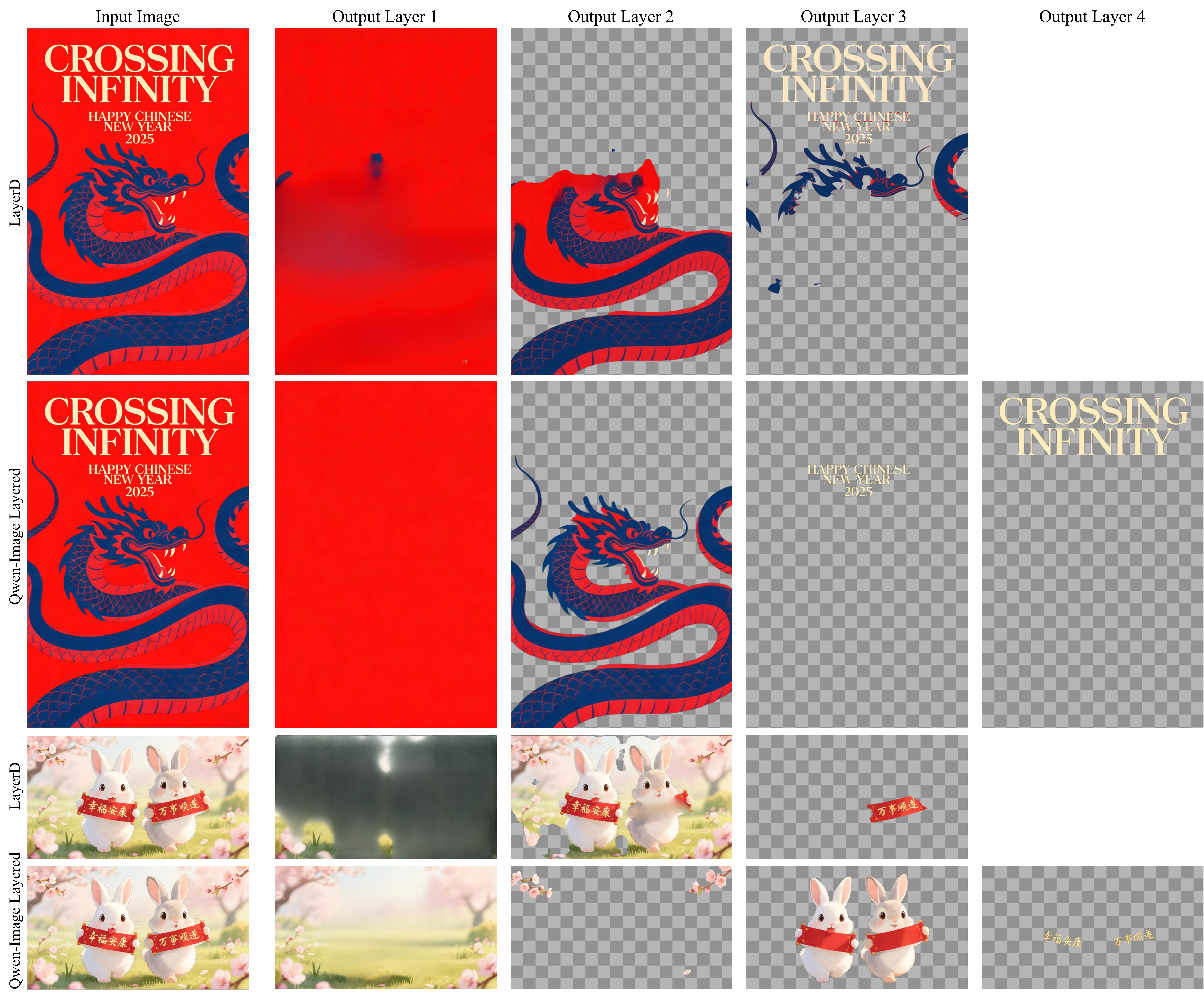}
    \vspace{-8mm}
    \caption{Qualitative comparison of Image-to-Multi-RGBA (I2L). The leftmost column shows the input image; the subsequent columns present the decomposed layers. Notably, LayerD~\cite{suzuki2025layerd} exhibits inpainting artifacts (Output Layer 1) and inaccurate segmentation (Output Layer 2 and 3), while our method produces high-quality, semantically disentangled layers, suitable for inherently consistent image editing. }
    \label{fig:qualitative}
    \vspace{-2mm}
\end{figure*}

\begin{table*}[!ht]
\centering
\caption{Quantitative comparison of Image-to-Multi-RGBA (I2L) on Crello dataset~\cite{yamaguchi2021canvasvae}. RGB L1: L1 distance between RGB channels weighted by the ground-truth alpha. Alpha soft IoU: soft IoU between predicted and ground-truth alpha channel.}\label{tab:quantitative}
\vspace{-2mm}
\begin{adjustbox}{width=\textwidth}
\begin{tabular}{l|cccccc|cccccc}
\toprule
Metric                 & \multicolumn{6}{c|}{RGB L1$\downarrow$}                          & \multicolumn{6}{c}{Alpha soft IoU$\uparrow$}  
\\
\midrule
\# Max Allowed Layer Merge               & 0      & 1      & 2      & 3      & 4      & 5      & 0      & 1      & 2      & 3      & 4      & 5      \\
\midrule
VLM Base + Hi-SAM ~\cite{chen2025rethinking}     & 0.1197 & 0.1029 & 0.0892 & 0.0807 & 0.0755 & 0.0726 & 0.5596 & 0.6302 & 0.6860 & 0.7222 & 0.7465 & 0.7589 \\
Yolo Base + Hi-SAM     & 0.0962 & 0.0833 & 0.0710 & 0.0630 & 0.0592 & 0.0579 & 0.5697 & 0.6537 & 0.7169 & 0.7567 & 0.7811 & 0.7897 \\
LayerD~\cite{suzuki2025layerd}                 & 0.0709 & 0.0541 & 0.0457 & 0.0419 & 0.0403 & 0.0396 & 0.7520 & 0.8111 & 0.8435 & 0.8564 & 0.8622 & 0.8650 \\
\midrule
\textbf{Qwen-Image-Layered-I2L} & \textbf{0.0594} & \textbf{0.0490} & \textbf{0.0393} & \textbf{0.0377} & \textbf{0.0364} & \textbf{0.0363} & \textbf{0.8705} & \textbf{0.8863} & \textbf{0.9105} & \textbf{0.9121} & \textbf{0.9156} & \textbf{0.9160} \\
\bottomrule
\end{tabular}
\end{adjustbox}
\vspace{-5mm}
\end{table*}

\subsection{Implementation Details}
Building upon Qwen-Image~\cite{wu2025qwen}, we developed Qwen-Image-Layered. The model was trained using the Adam optimizer~\cite{adam2014method} with a learning rate of $1 \times 10^{-5}$. For Text-to-RGB and Text-to-RGBA generation, training was performed on an internal dataset. For both Text-to-Multi-RGBA and Image-to-Multi-RGBA generation, the model was optimized on our proposed multilayer image dataset, with the maximum number of layers set to 20. The training process was conducted in three stages, comprising 500K, 400K, and 400K optimization steps, respectively.

\subsection{Quantitative Results}
\begin{figure*}[!h]
    \centering
    \includegraphics[width=\linewidth]{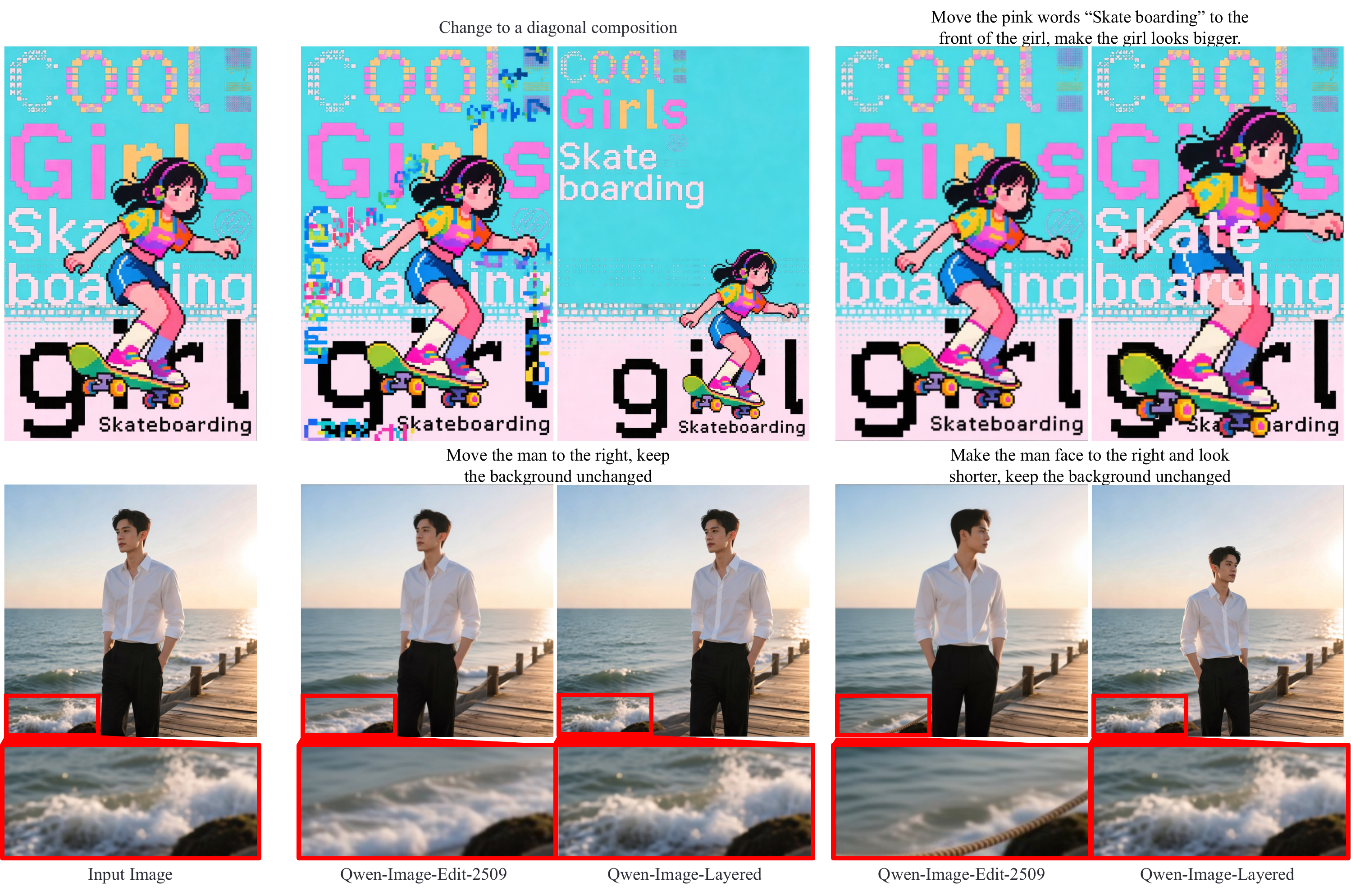}
    \vspace{-6mm}
    \caption{Qualitative comparison of image editing. The leftmost column is the input image; prompts are listed above each row. Qwen-Image-Edit-2509~\cite{wu2025qwen} struggles with resizing and repositioning, tasks inherently supported by Qwen-Image-Layered. Meanwhile, Qwen-Image-Edit-2509 introduces pixel-level shifts (last row), while Qwen-Image-Layered can ensure consistency by editing specific layers.}
    \label{fig:editing.}
    \vspace{-2mm}
\end{figure*}

\subsubsection{Image Decomposition}\label{sec:decomposition}
To quantitatively evaluate image decomposition, we adopt the evaluation protocol introduced by LayerD~\cite{suzuki2025layerd}. This protocol aligns layer sequences of varying lengths using order-aware Dynamic Time Warping and allows for the merging of adjacent layers to account for inherent ambiguities in decomposition (i.e., a single image may have multiple plausible decompositions). Quantitative results on Crello dataset~\cite{yamaguchi2021canvasvae} are reported in \cref{tab:quantitative}. Following LayerD~\cite{suzuki2025layerd}, we report two metrics: RGB L1 (the L1 distance of the RGB channels weighted by the ground-truth alpha) and Alpha soft IoU (the soft IoU between predicted and ground-truth alpha channels). Due to a significant distribution gap between the Crello dataset and our proposed multilayer dataset—such as differences in the number of layers and the presence of semi-transparent layers—we finetune our model on Crello training set. As shown in \cref{tab:quantitative}, our method achieves the highest decomposition accuracy, notably achieving a significantly higher Alpha soft IoU score, underscoring its superior ability in generating high-fidelity alpha channels.

\begin{table*}[!ht]
\centering
\caption{Ablation study on Crello dataset~\cite{yamaguchi2021canvasvae}. L: Layer3D Rope, R: RGBA-VAE, M: Multi-stage Training.}\label{tab:ablation}
\vspace{-2mm}
\begin{adjustbox}{width=\textwidth}
\begin{tabular}{
  >{\arraybackslash}p{5.5cm}|
  >{\centering\arraybackslash}p{0.2cm}
  >{\centering\arraybackslash}p{0.2cm}
  >{\centering\arraybackslash}p{0.2cm}|
  cccccc|
  cccccc}
\toprule
Metric    & \multicolumn{3}{c|}{Component}           & \multicolumn{6}{c|}{RGB L1$\downarrow$}                          & \multicolumn{6}{c}{Alpha soft IoU$\uparrow$}  
\\
\midrule
\# Max Allowed Layer Merge   & L & R & M           & 0      & 1      & 2      & 3      & 4      & 5      & 0      & 1      & 2      & 3      & 4      & 5      \\
\midrule
\textbf{Qwen-Image-Layered-I2L-w/o LRM} & $\times$ & $\times$  & $\times$ & 0.2809 & 0.2567 & 0.2467 & 0.2449 & 0.2439 & 0.2435 & 0.3725 & 0.4540 & 0.5281 & 0.5746 & 0.5957 & 0.6031 \\
\textbf{Qwen-Image-Layered-I2L-w/o RM} & \checkmark & $\times$ & $\times$ & 0.1894 & 0.1430 & 0.1255 & 0.1173 & 0.1138 & 0.1126 & 0.5844 & 0.6927 & 0.7576 & 0.7847 & 0.7954 & 0.7984 \\
\textbf{Qwen-Image-Layered-I2L-w/o M} & \checkmark & \checkmark & $\times$ & 0.1649 & 0.1178 & 0.1048 & 0.0992 & 0.0966 & 0.0959 & 0.6504 & 0.7583 & 0.8074 & 0.8243 & 0.8310 & 0.8331 \\
\midrule
\textbf{Qwen-Image-Layered-I2L} & \checkmark & \checkmark & \checkmark & \textbf{0.0594} & \textbf{0.0490} & \textbf{0.0393} & \textbf{0.0377} & \textbf{0.0364} & \textbf{0.0363} & \textbf{0.8705} & \textbf{0.8863} & \textbf{0.9105} & \textbf{0.9121} & \textbf{0.9156} & \textbf{0.9160} \\
\bottomrule
\end{tabular}
\end{adjustbox}
\vspace{-4mm}
\end{table*}

\begin{table}[htbp]
\centering
\caption{Quantitative comparison of RGBA image reconstruction on the AIM-500 dataset~\cite{li2021deep}. }\label{tab:vae}
\vspace{-2mm}
\begin{adjustbox}{width=0.48\textwidth}
\begin{tabular}{l|l|
>{\centering\arraybackslash}m{1cm}
>{\centering\arraybackslash}m{1cm}
>{\centering\arraybackslash}m{1cm}
>{\centering\arraybackslash}m{1cm}
}
\toprule
Model &   Base Model         & PSNR$\uparrow$    & SSIM$\uparrow$ & rFID$\downarrow$    & LPIPS$\downarrow$  \\
\midrule
LayerDiffuse~\cite{zhang2024transparent}              & SDXL       & 32.0879 & 0.9436 & 17.7023 & 0.0418 \\
\multirow{2}{*}{AlphaVAE~\cite{wang2025alphavae}} & SDXl       & 35.7446 & 0.9576 & 10.9178 & 0.0495 \\
                          & FLUX       & 36.9439 & 0.9737 & 11.7884 & 0.0283 \\
\midrule
\textbf{RGBA-VAE}                    & Qwen-Image & \textbf{38.8252} & \textbf{0.9802} & \textbf{5.3132}  & \textbf{0.0123} \\
\bottomrule
\end{tabular}
\end{adjustbox}
\vspace{-6mm}
\end{table}

\begin{figure*}[!h]
    \centering
    \includegraphics[width=\linewidth]{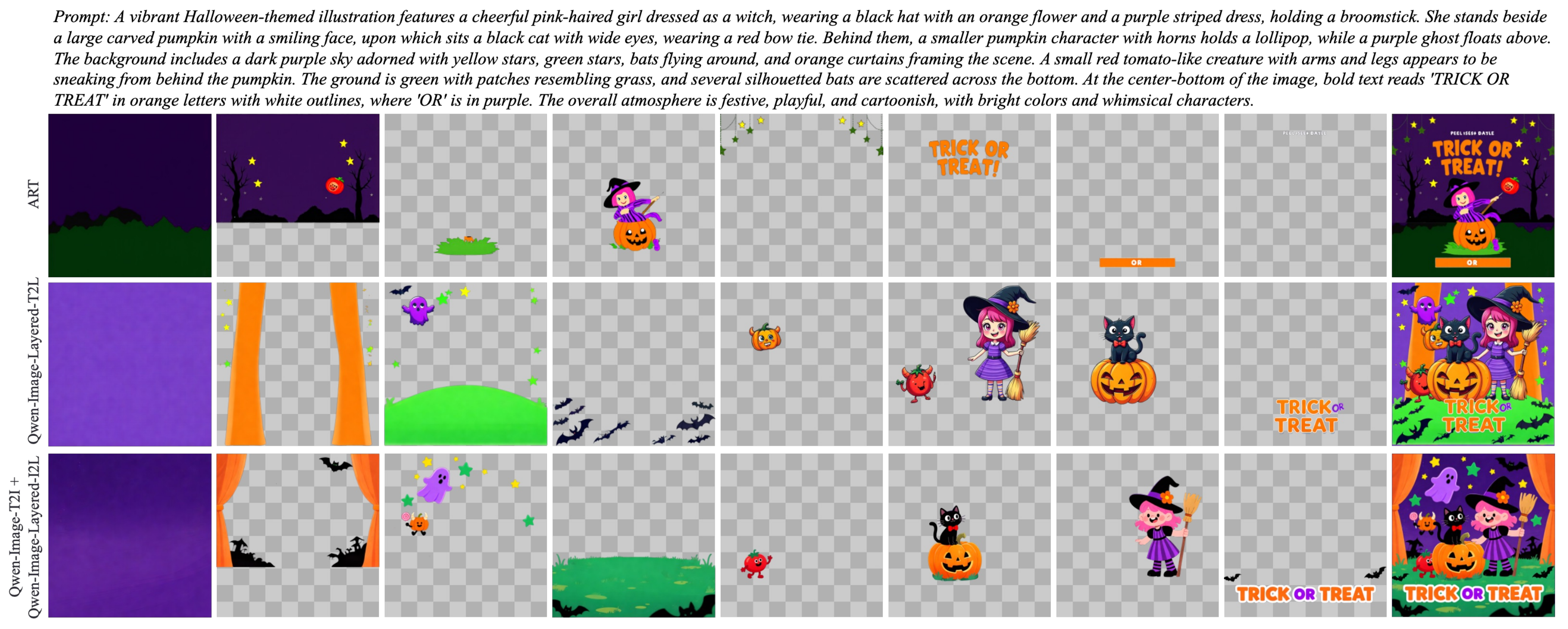}
    \vspace{-8mm}
    \caption{Qualitative comparison of Text-to-Multi-RGBA (T2L). The rightmost column shows the composite image. The second row directly generates layers from text (Qwen-Image-Layered-T2L); the third row first generates a raster image (Qwen-Image-T2I) then decomposes it into layers (Qwen-Image-Layered-I2L). ART~\cite{pu2025art} fails to follow the prompt, while Qwen-Image-Layered-T2L produces semantically coherent layers, and Qwen-Image-T2I + Qwen-Image-Layered-I2L further improves visual aesthetics.}
    \vspace{-6mm}
    \label{fig:mli}
\end{figure*}
\subsubsection{Ablation Study}

We conducted an ablation study on Crello dataset~\cite{yamaguchi2021canvasvae} to validate the effectiveness of our proposed method. The results are presented in \cref{tab:ablation}. For settings without multi-stage training, we initialize the model directly from pretrained text-to-image weights. For experiments without RGBA-VAE, we employ the original RGB VAE to encode the input RGB image while retaining RGBA-VAE for output RGBA layers. For variants without Layer3D RoPE, we replace it with standard 2D RoPE for positional encoding. All ablation experiments follow the same evaluation protocol as described in \cref{sec:decomposition}. As shown in the third and fourth rows, multi-stage training effectively improves decomposition quality. Comparing the second and third rows, the superior performance in the third row indicates that RGBA VAE effectively eliminates the distribution gap, thereby improving overall performance. Furthermore, the comparison between the first and second rows illustrates the necessity of Layer3D Rope: without it, the model can not distinguish between different layers, thus failing to decompose images into multiple meaningful layers.

\subsubsection{RGBA Image Reconstruction}
Following AlphaVAE~\cite{wang2025alphavae}, we quantitatively evaluate RGBA image reconstruction by blending the reconstructed images over a solid-color background. Quantitative results on AIM-500 dataset~\cite{li2021deep} are presented in \cref{tab:vae}, where we compare our proposed RGBA VAE against LayerDiffuse~\cite{zhang2024transparent} and AlphaVAE~\cite{wang2025alphavae} in terms of PSNR, SSIM, rFID, and LPIPS. As shown in \cref{tab:vae}, RGBA VAE achieves the highest scores across all four metrics, demonstrating its outstanding reconstruction capability.

\subsection{Qualitative Results}

\subsubsection{Image Decomposition}
We present a qualitative comparison of image decomposition with LayerD~\cite{suzuki2025layerd} in \cref{fig:qualitative}. Notably, LayerD produces low-quality decomposition layers due to inaccurate segmentation (layers 2 and 3) and inpainting artifacts (layer 1), rendering its results unsuitable for editing. In contrast, our model performs image decomposition in an end-to-end manner without relying on external modules, yielding more coherent and semantically plausible decompositions, thereby facilitating inherently consistent image editing.

\subsubsection{Image Editing}
In \cref{fig:editing.}, we present a qualitative comparison with Qwen-Image-Edit-2509~\cite{wu2025qwen}. For Qwen-Image-Layered, we first decompose the input image into multiple semantically disentangled RGBA layers and then apply simple manual edits. As illustrated, Qwen-Image-Edit-2509 struggles to follow instructions involving layout modifications, resizing, or repositioning. In contrast, Qwen-Image-Layered inherently supports these elementary operations with high fidelity. Moreover, Qwen-Image-Edit-2509 introduces noticeable pixel-level shifts, as shown in the bottom row. By contrast, layered representation enables precise editing of individual layers while leaving others exactly untouched, thereby achieving consistency-preserving editing.

\subsubsection{Multilayer Image Synthesis}

In \cref{fig:mli}, we present a qualitative comparison of Text-to-Multi-RGBA generation. In the second row, we directly employ Qwen-Image-Layered-T2L for text-conditioned multilayer image synthesis. Alternatively, we first generate a raster image from text using Qwen-Image-T2I~\cite{wu2025qwen} and then decompose it into multiple layers using Qwen-Image-Layered-I2L. As illustrated, ART~\cite{pu2025art} struggles to generate semantically coherent multilayer images (\eg missing bats and cat). In contrast, Qwen-Image-Layered-T2L produces semantically coherent multilayer compositions. Moreover, the pipeline combining Qwen-Image-T2I and Qwen-Image-Layered-I2L further leverages the knowledge embedded in the text-to-image generator, enhancing both semantic alignment and visual aesthetics.

\section{Conclusion}
\label{sec:conclusion}

In this paper, we introduce Qwen-Image-Layered, an end-to-end diffusion model that decomposes a single RGB image into multiple semantically disentangled RGBA layers. By representing images as a stack of layers, our approach enables inherent editability: each layer can be independently manipulated while leaving all other content exactly unchanged, thereby fundamentally ensuring consistency across edits. Extensive experiments demonstrate that our method significantly outperforms existing approaches in decomposition quality and establishes a new paradigm for consistency-preserving image editing. 
{
    \small
    \bibliographystyle{ieeenat_fullname}
    \bibliography{main}
}

\end{document}